# Using Machine Learning to Predict Engineering Technology Students' Success with Computer Aided Design

Jasmine Singh[1], Viranga Perera*[1], Alejandra J. Magana[2], Brittany Newell[3], Jin Wei-Kocsis[2], Ying Ying Seah[4], Greg J. Strimel[5], and Charles Xie[6]

[1] Purdue Polytechnic Institute, Purdue University, West Lafayette, IN, USA.

[2] Department of Computer and Information Technology, Purdue University, West Lafayette, IN, USA.

[3] School of Engineering Technology, Purdue University, West Lafayette, IN, USA.

[4] School of Business, Oakland City University, Oakland City, IN, USA.

[5] Department of Technology Leadership & Innovation, Purdue University, West Lafayette, IN, USA.

[6] Institute for Future Intelligence, Natick, MA, USA.

*Corresponding author: viranga@purdue.edu

**Abstract**

Computer-aided design (CAD) programs are essential to engineering as they allow for better designs through low-cost iterations. While CAD programs are typically taught to undergraduate students as a job skill, such software can also help students learn engineering concepts. A current limitation of CAD programs (even those that are specifically designed for educational purposes) is that they are not capable of providing automated real-time help to students. To encourage CAD programs to build in assistance to students, we used data generated from students using a free, open source CAD software called *Aladdin* to demonstrate how student data combined with machine learning techniques can predict how well a particular student will perform in a design task. We challenged students to design a house that consumed zero net energy as part of an introductory engineering technology undergraduate course. Using data from 128 students, along with the *scikit-learn* Python machine learning library, we tested our models using both total counts of design actions and sequences of design actions as inputs. We found that our models using early design sequence actions are particularly valuable for prediction. Our logistic regression model achieved a >60% chance of predicting if a student would succeed in designing a zero net energy house. Our results suggest that it would be feasible for *Aladdin* to provide useful feedback to students when they are approximately halfway through their design. Further improvements to these models could lead to earlier predictions and thus provide students feedback sooner to enhance their learning.

**1.0 Introduction**

*Engineering design* is a complex, interactive problem-solving process of satisfying a set of specific constraints using scientific and engineering knowledge [15]. Sheppard (2003) characterized *engineering design* as a method to "scope, generate, evaluate, and realize ideas" [18]. This process requires experience and understanding of multiple disciplines to address real-world problems. Consequently, it is a challenging concept to teach undergraduate students [8; 12], particularly first- and second-year students who are in the process of learning fundamental disciplinary concepts and have limited experience in the field. As a further complication, the Fourth Industrial Revolution (i.e., Industry 4.0) and the Internet of Things require virtual prototyping and integration between the physical and the digital worlds. As such, Computer-Aided Design (CAD) software is increasingly becoming both a central tool of the modern engineering design process as well as a critical educational tool.

While the engineering industry needs incoming employees to be skilled in using CAD software [20; 21], CAD software can also be an essential tool for teaching the engineering design process to undergraduate students. In general, CAD software assists the "creation, modification, analysis, or optimization of a design" [9]. As for education, CAD software offers flexibility for students to create and test their designs in various configurations without the need for physical facilities or equipment, which is safer and more cost effective [7; 23]. Using CAD solutions can enable students to model mechanical and mechatronic systems using a base geometry and parametric representations eliminating physical constraints while allowing students to build and see product and process interactions [5].

Modern CAD software also provides an additional benefit for education by allowing researchers to collect user interactions in an unobstructive manner, so that we can study how individual students go through the design process [22]. This allows us to observe, characterize, and as we demonstrate in this work, predict students' design performance. Research in this area can provide technical solutions that help instructors who may not be able to attend to every student's needs in real-time. Compared with other disciplines in which data mining is viewed as a way to develop instructional intelligence, engineering design—a highly open-ended cognitive and creative process that is not well understood—may need this kind of automatic tools even more. As there is no single correct answer to a design problem, every idea may need to be taken seriously and evaluated objectively. This exerts heavy burdens on instructors and calls for the assistance of machine learning algorithms [13].

In this work, we used data generated by undergraduate students using *Aladdin*, a CAD software program developed by the Institute for Future Intelligence, during an introductory engineering technology course. Students were tasked with designing a house that consumed zero net energy using *Aladdin*. As they used *Aladdin*, their design actions were logged into JavaScript Object Notation (JSON) files. We used those files along with a machine learning technique to create a Python code that can predict if randomly selected students would achieve the design goal. With this work, we aim to answer the following research questions: (1). How does this work help with implementing automated CAD interventions to help future learners? (2). How does this work support design and manufacturing within Industry 4.0?

## 2.0 Methods

### 2.1 Course Details

Student data for this work came from the ENGT 18000 (Engineering Technology Foundations) course at Purdue University that took place during the Fall 2020 semester. Students who enroll in this course are typically first-year undergraduate students from the School of Engineering Technology. Due to the COVID-19 pandemic, students had the option of either taking the course in a hybrid format (i.e., where the instructor delivered lectures online, but students were expected to attend recitations in person) or in a distance learning format (i.e., where all components of the course were conducted online). During the Fall 2020 semester, there were a total of 323 students enrolled (248 students in the hybrid section and 75 students in the distance learning section).

The course covered a variety of introductory topics including plotting, programing and data analysis using Excel and MATLAB, conducting experiments and reporting results, basic statistics, energy, series and parallel circuits, and statics (mechanics). For this work, we focused on the energy module of the course, which spanned two weeks of the semester. The energy module consisted of a pre-quiz that gauged students' basic understanding of energy concepts, a lecture, two activities based on *Aladdin* (discussed in more detail in Section 2.2), and a post-quiz that again gauged students' basic understanding of energy concepts. The lecture covered different forms of energy (e.g., potential, kinetic, and internal), transformation of energy, the first and second laws of thermodynamics, work, and power. The energy section of the course was also used as a bridge to help students understand conversions between electrical and mechanical concepts taught in the course.

### *2.2 Aladdin CAD Software*

To help students learn energy concepts during the course, we used a free, open-source CAD software tool called *Aladdin* (formerly known as *Energy3D*) [23]. *Aladdin* has been specifically designed to help students from middle school through college learn engineering design through computer modeling and simulation. Its intuitive and easy to use interface allows students to quickly design 3D buildings (see Figure 1 for an example CAD house designed in *Aladdin*) and to then conduct iterative analyses to improve their design step by step. Features of *Aladdin* include adjusting dimensions of a building, changing components of a building (e.g., doors and windows), including trees for shade, and adding solar panels. In addition to design features, *Aladdin* also includes analysis tools such as determining the path of the sun throughout the year as a function of geographic location and calculating the net energy use of a building throughout the year.

### *2.3 Course Implementation*

As mentioned in Section 2.1, we had students use *Aladdin* as part of the energy module of the course. Students were given instructions on how to install the software on their own computers and were directed to YouTube tutorials that demonstrated how to use the software. Two researchers also went to the in-person recitation sessions to answer questions and help students with any issues they had with the software. While this additional help was not available to the distance section, we provided answers to frequently asked questions at the recitation sections through the learning management system to those students. The instructor and two researchers also answered emails from students and provided answers in online meetings directly with students.

The first activity involving *Aladdin* was a small design exercise that was intended to help students get familiar with the software. For the small design exercise, students were asked to build an energy-efficient house using *Aladdin*. They were asked to choose one factor to change (e.g., solar panel tilt), make a prediction of what would happen from the change by providing reasoning for what they thought would happen due to the change (using the Claim, Evidence, and Reasoning [CER] framework [24]), then observe what happened due to the change, and finally justify why they thought the change happened (again using the CER framework). See Supporting Information for the small design exercise journal that students had to complete.

The second activity with *Aladdin* was a more involved design challenge. We had students design a zero net energy house (viz., a house that consumed zero net energy throughout the year). They were asked to design the house in Indianapolis, Indiana (being the closest city available within *Aladdin* that was near the Purdue University campus). They were also instructed that the house should not cost more than $200,000, that it should comfortably fit a four-person family (i.e., have an area between 150 and 200 square meters and a total height [with wall and roof] of between 6 to 10 meters). Additionally, each side of the house needed to have at least one window, tree trunks needed to be at least 2 meters away from the house, and solar panels could not hang over roof edges. Similar to the small design exercise, students were asked to document their design process by making predictions, observing changes, and justifying reasons for those changes in a design journal. See Supporting Information for the design challenge journal that students had to complete.

### 2.4 Aladdin Data

*Aladdin* records actions that users take within the software into JSON files. Those JSON files contain individual actions (e.g., adding a wall and moving a solar panel) with timestamps, which provide great detail to retrace students' steps to better understand their design process. See Supporting Information for an example JSON file from the course.

The design challenge was selected as the focus for this study since the primary reason for the small design exercise was to get students familiar with the software. We first cleaned the JSON files by correcting formatting issues and removing empty files. Empty files were eliminated by traversing through all files and deleting any files with zero size. Occasionally, *Aladdin* produces incorrectly formatted JSON files, for example, due to a user abruptly closing the program. To account for files with errors, we created a JSON parsing tool to print out and highlight specific errors in each file (see Supporting Information for a link to the Python code). Many corrections involved adding a missing punctuation mark or removing non-UTF-8 characters. While the total number of students enrolled in both sections of the course was 323, the total number of students with complete, error-corrected JSON files for their design challenge was 128. As such, those 128 students served as the cohort for this study.

### 2.5 Machine Learning Model

For our machine learning models, we used the open-source *scikit-learn* Python library [16]. For each student, we used two types of inputs for our models: (1) tallied design actions and (2) sequence of design actions. The tallied design actions served as a preliminary test for our modeling, with the sequence of design actions being the input that would enable predictions for the purpose of providing real-time feedback to students in the future. Additionally, we used both linear and logistic regression models when trying to predict final net energy values. The linear regression served as a preliminary test for our modeling, with the logistic regression being the primary focus of our results. We trained all our models using an 80-20 data split (i.e., 80% of the data used for training and 20% used for prediction), which is a common practice in machine learning work [e.g., 17].

In the case of the tallied design actions, we summed design actions (e.g., adding, removing, and editing a building element) by each category (e.g., wall, roof, and solar panel) for each student. Those tallied actions were then added to a *Pandas* (a Python data analysis library) data frame where data for each student was a row and each tallied action was a column. Since the goal of the design project was to design a zero net energy house, having a zero net energy value at the end was considered to be the metric of success. The final net energy for each student was also added to a separate *Pandas* data frame where each student was a row and the column was their final net energy value. Students who did not have a final net energy value in their JSON data were removed since they could potentially skew the model. For our first set of models, for each student, their tallied design actions were the independent variables and their final net energy value was the dependent variable. We explored using both linear and logistic regression models (see results in Sections 3.1 and 3.2).

For the case of using sequences of design actions, for each student, we created arrays of their design actions and numerically coded each action based on the category of the action. For example, all actions pertaining to doors (e.g., adding, removing, or editing them) were coded as 0 and actions pertaining to solar panels were coded as 6 (see Table 1 for a complete list of codes). Those numerically coded sequences of actions then served as independent variables for our models (with the dependent variable still being the final net energy values). For these models, we only used logistic regression models (see results in Section 3.3) since it was more important for the model to be able to predict if students were approximately close to a zero net energy house design rather than having the model try to predict the precise final net energy value (as is the case with a linear regression) for each student.

**3.0 Results**

***3.1 Linear Regression Model Using Tallied Design Actions***

In Figure 2 we show a histogram of the final net energies for our cohort. The histogram shows that most students successfully designed a zero net energy house. Figure 3 compares actual and predicted final net energies for 11 randomly selected students (20% of 55 students) for our initial tests of our linear regression model with tallied design actions. The figure demonstrates that a linear regression model is not capable of making accurate predictions based solely on tallied designed actions, which is not surprising since the model is attempting to determine the precise final net energy values for each student. For the purpose of helping a student with their design in real time, it is not necessary for a machine learning model to precisely predict their exact final net energy value. Instead the model should determine if a particular student is or is not progressing towards a successful design. As such, we decided that a logistic regression model would be better suited since we could set a certain range of final net energy values as being sufficiently close to the zero net energy design constraint.

### 3.2 Logistic Regression Model Using Design Action Counts

Since logistic regression models predict binary outcomes, we needed to choose the model dependent variable. We could have limited our cases to only students who designed a house that precisely used zero net energy. However, we determined that to be too restrictive since it likely would have removed students whose designs were close to zero net energy. To test what range of values near zero net energy to consider as meeting the design requirement, we ran logistic regression models for a series of final net energy ranges centered on 0 kWh. Figure 4 shows logistic regression model accuracies as a function of choosing various ranges of final net energy values to be considered as being within the design range. Based on these results, we decided that being within the range of -10,000 to +10,000 kWh meant achieving a zero net energy design. In Figure 5 we show results of one logistic regression model that shows how many final net energies it was able to predict correctly. Of course, due to the stochastic nature of regression models, we also tested our logistic regression model for stability. In Figure 6 we show accuracies of our logistic regression model for 10 iterations. The figure demonstrates that our model accuracy varies as expected but is consistently above 50% (with most cases being above 65%).

### 3.3 Logistic Regression Model Using Sequence of Design Actions

While our results in Sections 3.1 and 3.2 demonstrate that our regression models (particularly our logistic regression models) are capable of predicting student success based on their tallied design counts, they are not ideal for the purposes of providing students feedback in real time since they rely on tallied design actions, which are of course only available after students have completed their designs. As such, it is important to consider sequences of design actions as independent variables for our logistic regression models. In Figure 7 we show accuracies of our logistic regression models as a function of using varying percentages of the initial set of design action sequences. For example, using 10% of a student's action sequence means using only the first 10% of their design actions as the independent variable for our models. To demonstrate stability, we ran each percentage of action sequence for 10 iterations. Our results show that even when using only the very beginning of students' design action sequences (i.e., 10%), our models are generally better than chance at predicting if a particular student will successfully design a zero net energy house. If 60% of design action sequences are used, then our models have a 60% or greater chance of predicting success.

## 4.0 Discussion

### 4.1 How Does This Work Help With Implementing Automated CAD Interventions to Help Future Learners?

Providing students with feedback throughout their learning process is critical for their growth and success [6]. In the case of *design-based learning*, where students need to both learn and apply designated concepts [10; 11], *formative assessments* (also called *responsive teaching*) are especially necessary to provide students with timely feedback. This is the case since *design-based learning* has challenges such as "chance of design" whereas students may develop a successful solution without an in-depth understanding of the relevant theoretical concepts [1; 4; 14]. Specifically, when we asked students to design a zero net energy house, it is possible that students got to the desired outcome merely by trying a series of haphazard design actions. Thus, if not considered with caution, design-based learning challenges may limit a student's ability to transfer knowledge to novel contexts or problem scenarios [2]. This then requires great attention from instructors toward formative assessments and providing quality feedback to students. But, as any instructor knows, this is a time intensive process [3], and as such, is difficult to scale to all students especially in large classes and in the online learning environments that will only continue to become more prevalent.

Results from this research may begin to open the potential to scale quality formative assessment and real-time feedback when using CAD software such as *Aladdin*. We developed an algorithm that could begin to automate the process of providing just-in-time learning to students to improve their designs as well as the understanding of important engineering concepts related to these improvements. This type information can be tied to on-screen avatars in the future to engage learners in design dialogue, which guides them while they learn which design actions could lead to more successful outcomes and why. Stables (2017) piloted the use of digital avatars as surrogate mentors to design students in a way that prompted an ongoing conversation between the learner and the avatar [19]. These conversations were based on a dialogic framework that consisted of a sequence of questions that asked learners to a) describe what they were designing, b) provide an evaluation of their current progress, c) speculate on how their project could be improved, and d) to comment on their plans for next steps. Preliminary research on this approach indicated that the design dialog made students think more deeply about their design work. While this on-screen dialog was based on questions related to established design heuristics, this type of work could be tied to design efficiency data within the CAD program and, using machine learning, could provide just-in-time feedback to the students. By intervening and allowing students to recognize which of their actions will not likely lead to an optimal design, students can better learn what effect certain design actions have on outcomes such as the net energy use of a building. This would not only help students learn about design processes, but also support them in reaching success with the outputs of their design work. In summary, expanding this work has potential to scale the quality of feedback provided to students via formative assessments particularly for present and future online, hybrid, and HyFlex educational environments.

### 4.2 How Does This Work Support Design and Manufacturing Within Industry 4.0?

While further research related to this study and the implementation of machine learning within CAD programs can be positioned to enhance student learning, this work can also connect to, and support, the next generation of design and manufacturing. As design and manufacturing continue to converge to optimize and streamline production processes through online communities, the predictive nature of machine learning could provide opportunities to enhance the potential of design work. Today, artificial intelligence and machine learning are revolutionizing industry and providing artificial intelligence-driven parametric design, such as generative design, to quickly and easily generate thousands of variations of a design based on a set of desired criteria. This provides a large set of alternative directions for a design that one may have never considered. Although artificial intelligence may generate design options, there is still a need for decisions to be made when both inputting the criteria as well as selecting the best design for the situation. This is where machine learning algorithms can be of use in further supporting design work and increasing productivity while reducing time-consuming redesigns. The further implementation of these technologies can then also be used for learning within the workplace and human resource development. As such, CAD software will likely continue to implement machine learning and artificial intelligence to become a more critical tool for modern engineering design as well as education and workforce development.

## 5.0 Limitations and Future Work

There are several limitations to this study that can be improved with future work. One limitation is that we do not know why certain sequences of design actions produced successful zero net energy designs while others did not. In this work, we demonstrated that there are patterns in the design action sequences that help our machine learning model predict success. However, future work can explore the meanings of specific design actions, so that the machine learning model can be further improved. Another limitation is that we have not yet explored how students' design performance connected to their knowledge of energy concepts. Are better designers also the ones that have a better understanding of energy concepts? Are there some who performed poorly in the pre- and/or post-quizzes who performed well as designers? These are some of the questions that can be addressed in future work. Additionally, future work can improve the machine learning models by integrating additional emerging techniques such as recurrent neural networks (RNN).

## 6.0 Conclusions

We trained machine learning models to predict if a particular student would achieve a successful design as they used a CAD software to design a zero net energy house. This work indicates that it is possible to use students' design action sequences to predict their engineering design success. Specifically, our logistic regression model achieved a >60% chance of predicting if a student would succeed in designing a zero net energy house by using the first 60% of their design action sequences. While this means that we can currently implement interventions when students are about halfway through their design process, future improvements could lead to predictions of success based on a smaller set of design actions. That would in turn mean that we will be able to create software interventions earlier in the design process. Timely and effective feedback that is built into the CAD software can greatly improve student learning and engineering designs in the future.

## Supporting Information

We provide the small design exercise and the design challenge journals that students had to complete as Supporting Information. We also provide an example JSON file from the course. Codes developed as part of this work can be found at https://github.com/singh486/Aladdin_machine_learning

**Acknowledgments**

We thank the Purdue University Discovery Park Undergraduate Research Internship program for funding this work. We received approval for this work from Purdue University's Institutional Review Board (#IRB-2020-1294).

**Tables**

| Design Action Category | Description | Code |
|---|---|---|
| Door | Actions related to doors (e.g., add, edit, and remove) | 0 |
| Floor | Actions related to floors | 1 |
| Foundation | Actions related to the foundation | 2 |
| Wall | Actions related to walls | 3 |
| Window | Actions related to windows | 4 |
| Roof | Actions related to the roof | 5 |
| Solar panel | Actions related to solar panels | 6 |
| Tree | Actions related to trees | 7 |
| Building | Actions related to the whole building | 8 |
| Analysis | Actions related to analysis (e.g., show heliodon) | 9 |
| Parameters | Actions pertaining to geography (e.g., change latitude) | 10 |
| Thermal | Actions pertaining to thermal characteristics | 11 |
| Color | Actions that change colors of building components | 12 |

**Table 1**. Design action categories with brief descriptions for each along with their corresponding numerical codes

**Figures**

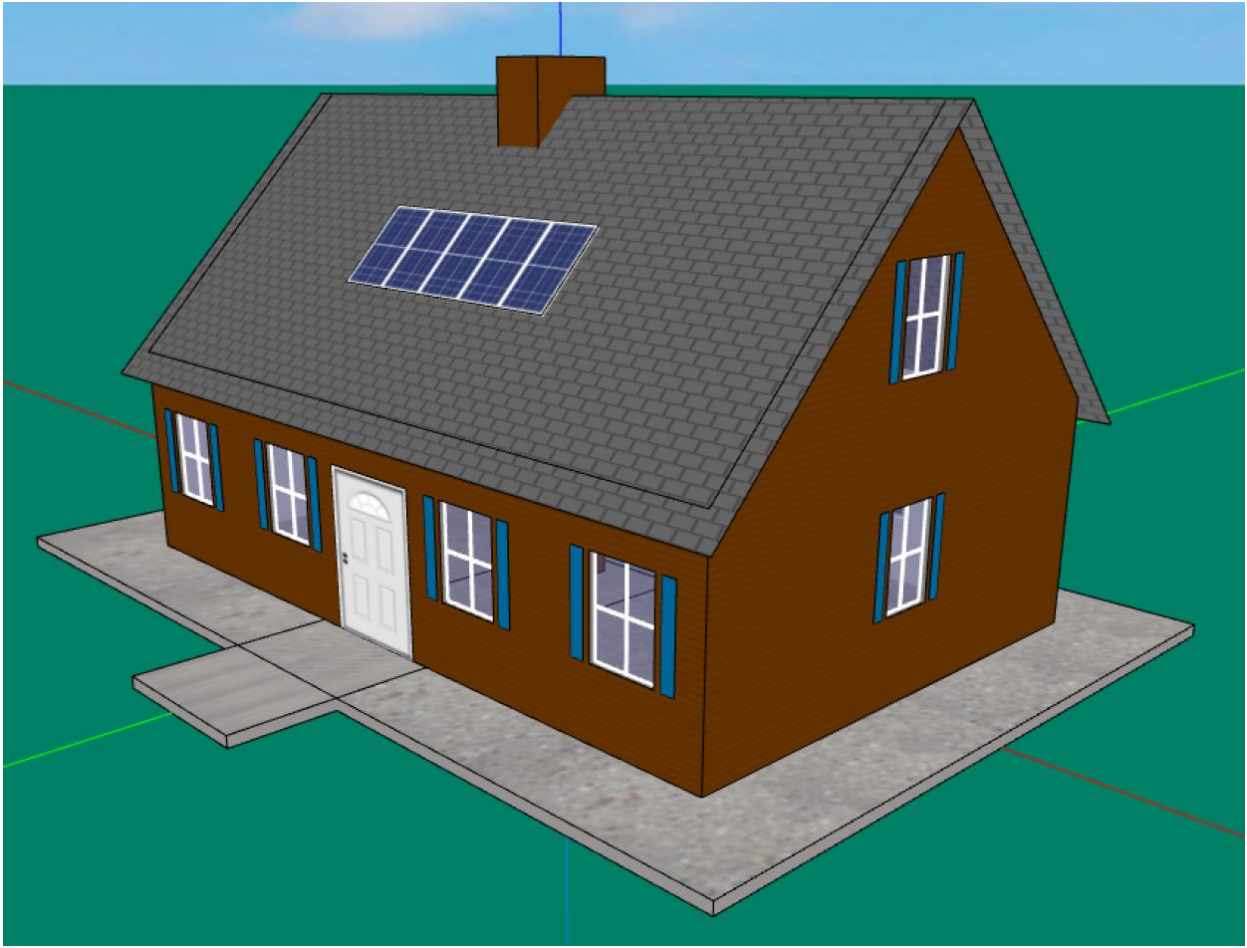

**Figure 1**. An example house in *Aladdin*

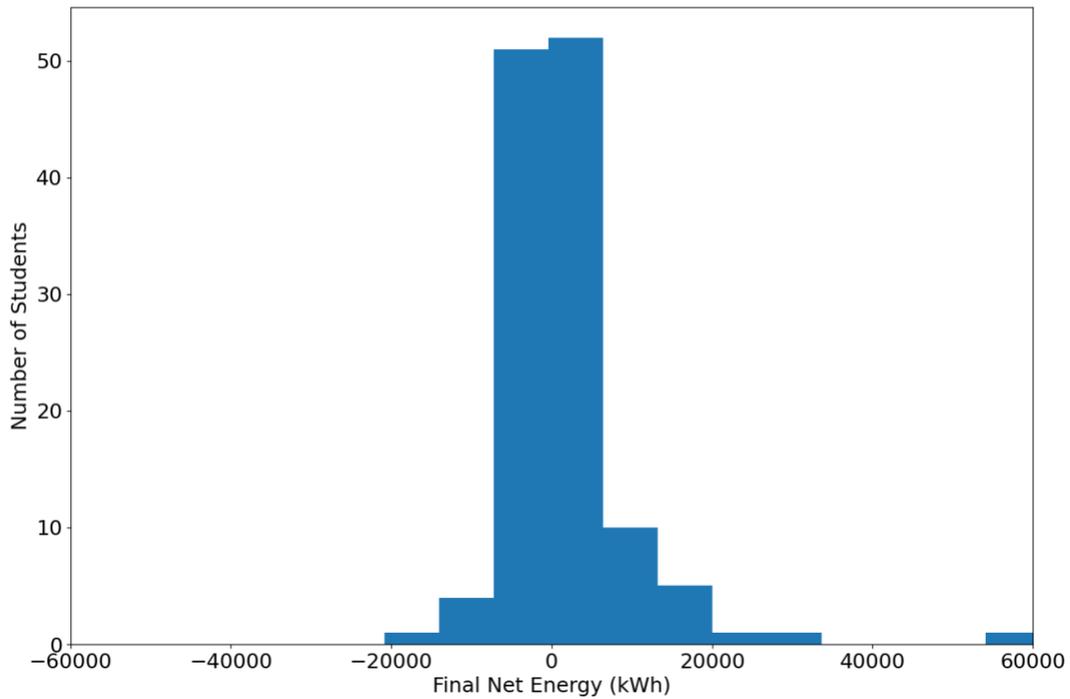

**Figure 2**. Histogram of final net energy values for the study cohort showing that most students achieved the goal of designing a zero net energy house. There are 2 outliers that do not appear on the figure (one student who had a final net energy of ~210,000 kWh and another who had ~660,000 kWh).

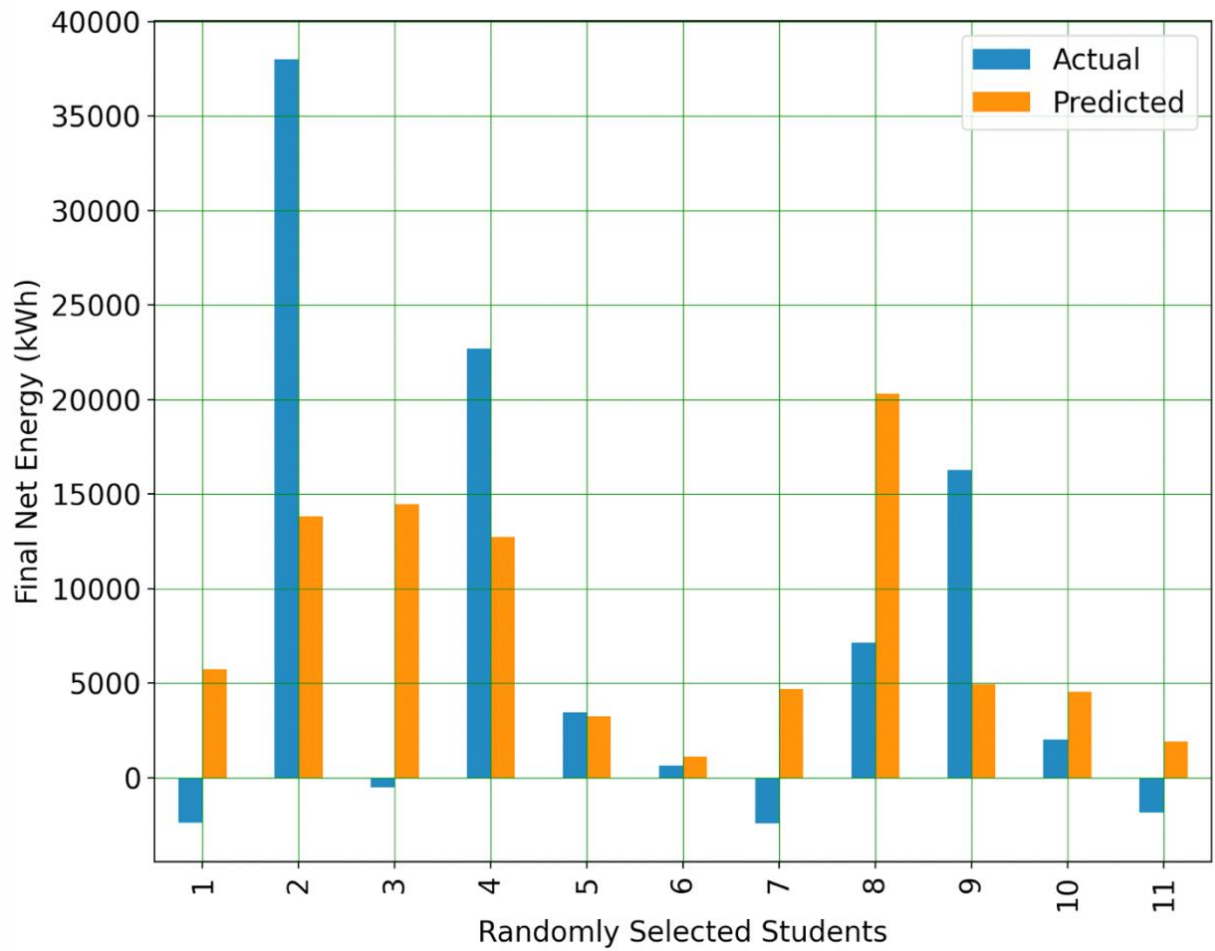

**Figure 3**. Actual (blue) and predicted (orange) final net energies for 11 randomly selected students using our linear model with tallied design action counts.

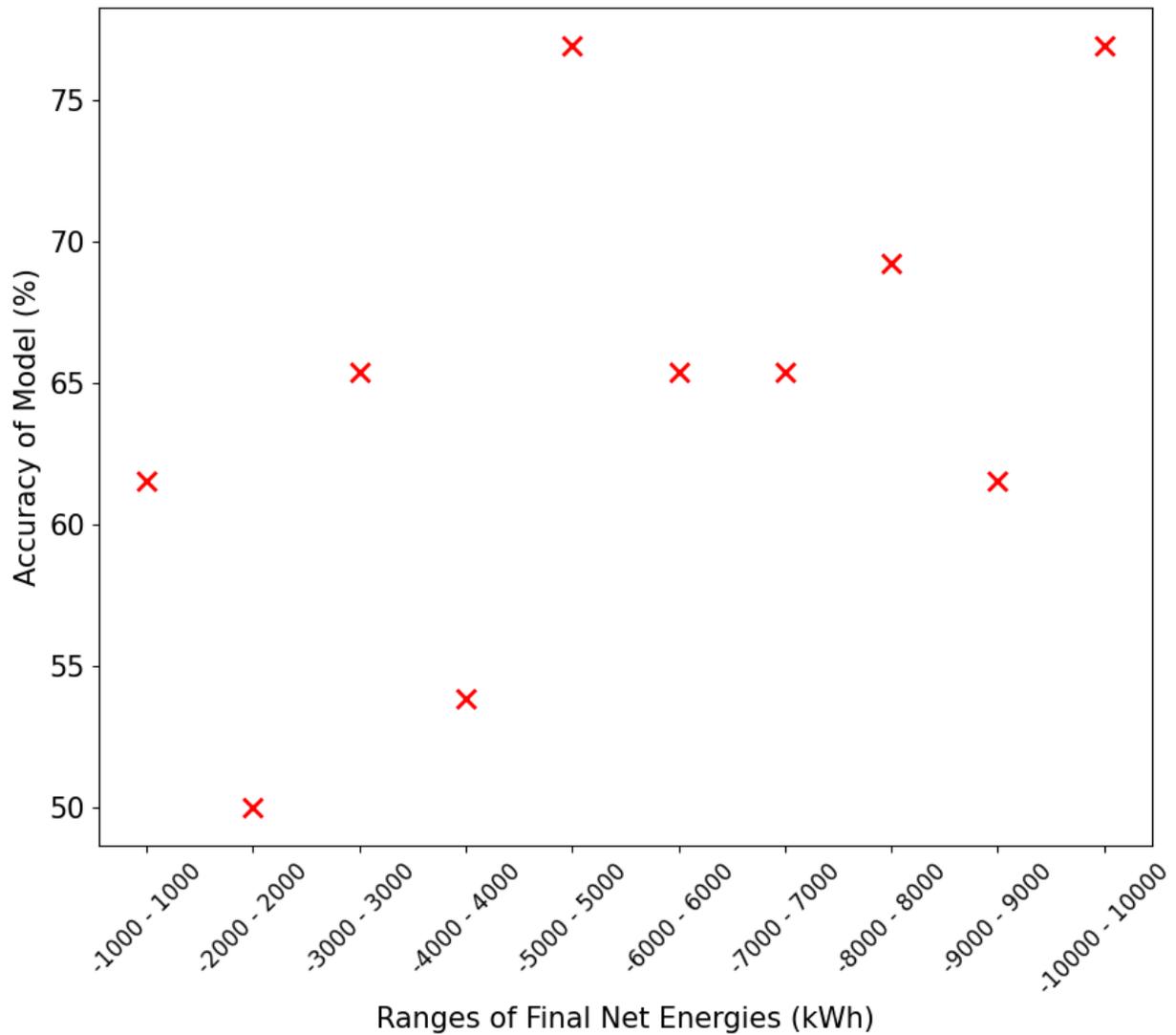

**Figure 4**. Logistic regression model accuracies as a function of choosing various ranges of final net energy values to be considered as being within the design range. Nominally we considered being within the range of -10,000 to +10,000 kWh as achieving a zero net energy design.

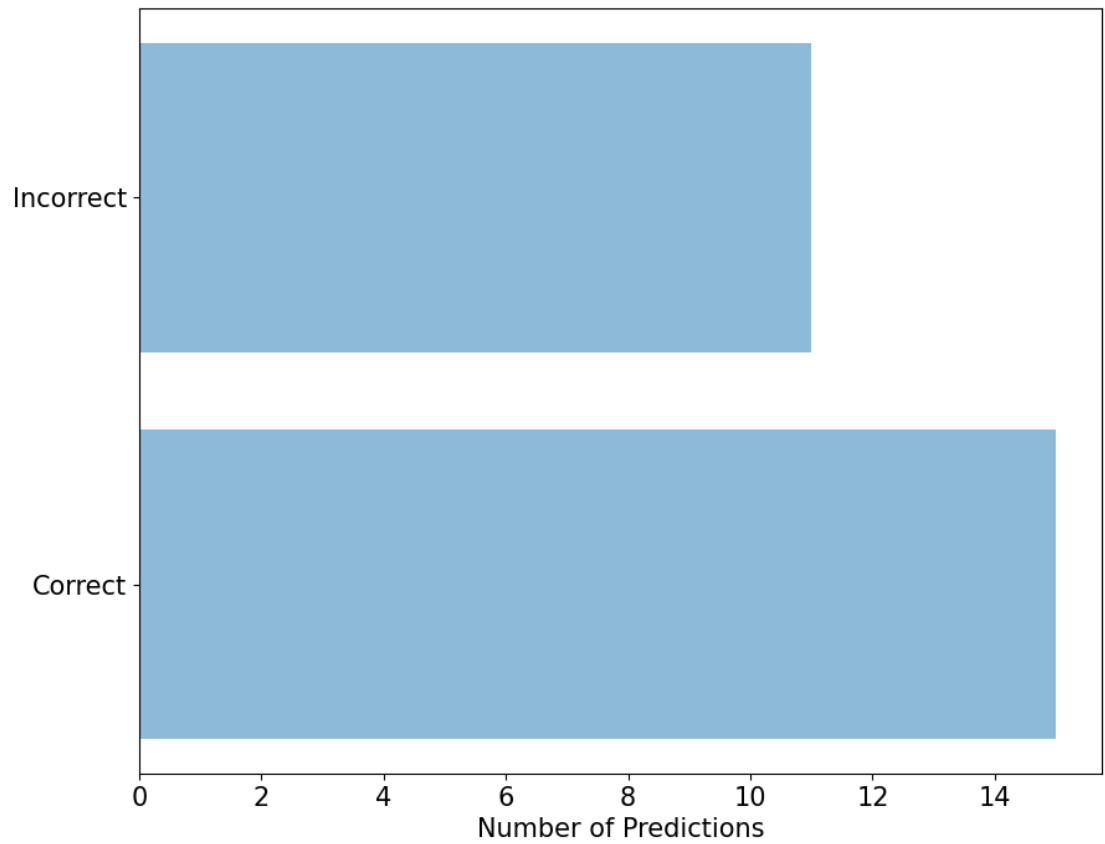

**Figure 5**. Results of a logistic regression model showing correct and incorrect predictions of achieving a zero net energy design using the complete sequence of design actions for each student.

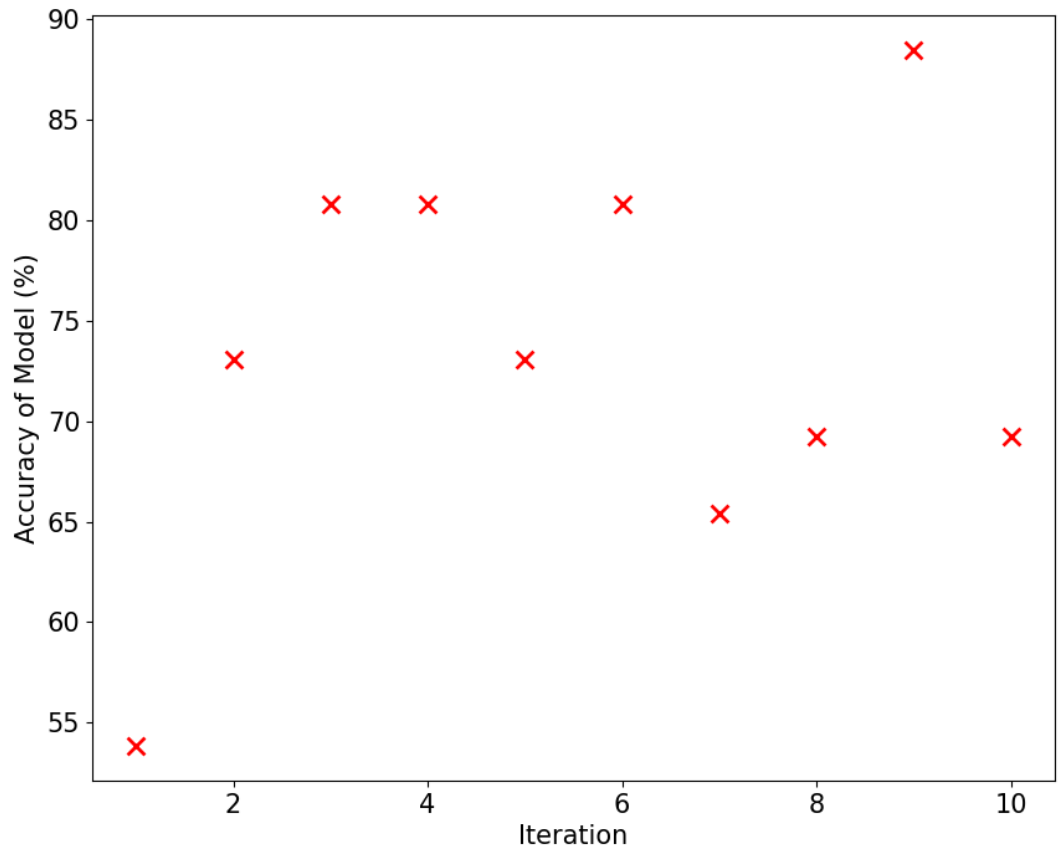

**Figure 6**. Accuracy of logistic regression models for 10 iterations to demonstrate that model accuracy varies as expected but is mostly above 65%.

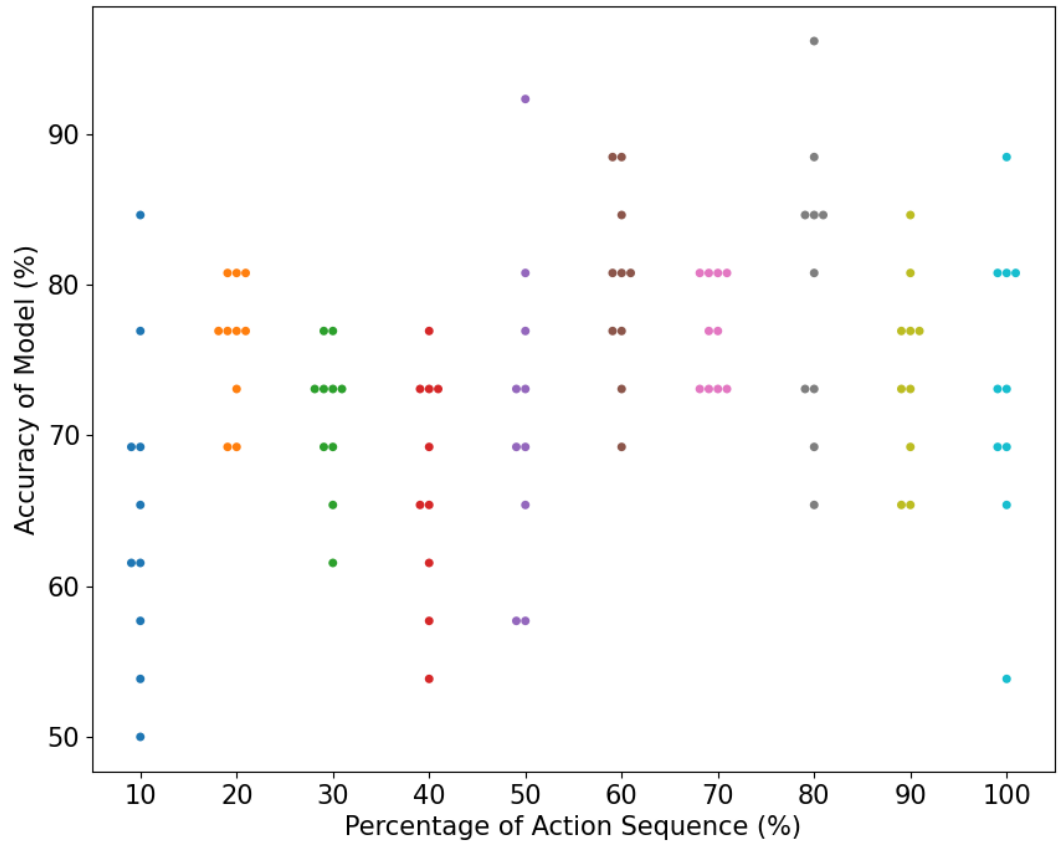

**Figure 7**. Accuracy of logistic regression models as a function of using varying percentages of the initial set of design action sequences (with 10% meaning using the first 10% of the design actions for each student and 100% meaning using the whole sequence of design actions). Each percentage of action sequence has been run for 10 iterations to demonstrate stability of the results. Repeated horizonal markers indicate multiple occurrences of the same accuracy for a given percentage of action sequence. Markers are colored by percentage of action of sequence for improved visibility.